\documentclass[journal,comsoc]{IEEEtran}

\usepackage[T1]{fontenc}% optional T1 font encoding

\usepackage{algorithm}
\usepackage{algorithmic}
\usepackage{lipsum}
\usepackage{tabularx,booktabs}
\usepackage{graphicx}
\usepackage{tikz}
\graphicspath{{images/}}
\usepackage{cite}
\usepackage{tabu}
\usepackage{hyperref}
\usepackage{array}
\usepackage{multirow}
\usepackage {mathtools}

\ifCLASSINFOpdf

\else

\fi

\usepackage{amsmath}

\usepackage[cmintegrals]{newtxmath}

\begin{document}

\title{Learning to Estimate 3D Human Pose \\ from Point Cloud}

\author{Yufan~Zhou,
        Haiwei~Dong,~\IEEEmembership{Senior Member,~IEEE}
        and~Abdulmotaleb~El~Saddik,~\IEEEmembership{Fellow,~IEEE}
        \thanks{Y. Zhou, H. Dong, and A. El Saddik are with the Multimedia Computing Research Laboratory, School of Electrical Engineering and Computer Science, University of Ottawa, Ottawa, ON K1N 6N5, Canada (email: yzhou158@uottawa.ca; haiwei.dong@ieee.org; elsaddik@uottawa.ca).}}
%\thanks{M. Shell was with the Department
%of Electrical and Computer Engineering, Georgia Institute of Technology, Atlanta,

%\markboth{Journal of \LaTeX\ Class Files,~Vol.~14, No.~8, August~2015}%{Shell \MakeLowercase{\textit{et al.}}: Bare Demo of IEEEtran.cls for IEEE Communications Society Journals}

\maketitle
\pagestyle{headings}

\begin{abstract}

3D pose estimation is a challenging problem in computer vision. Most of the existing neural-network-based approaches address color or depth images through convolution networks (CNNs). In this paper, we study the task of 3D human pose estimation from depth images. Different from the existing CNN-based human pose estimation method, we propose a deep human pose network for 3D pose estimation by taking the point cloud data as input data to model the surface of complex human structures. We first cast the 3D human pose estimation from 2D depth images to 3D point clouds and directly predict the 3D joint position. Our experiments on two public datasets show that our approach achieves higher accuracy than previous state-of-art methods. The reported results on both ITOP and EVAL datasets demonstrate the effectiveness of our method on the targeted tasks.

\end{abstract}

\IEEEpeerreviewmaketitle

\section{Introduction}

\IEEEPARstart{P}{ose} estimation aims to identify and locate the key points of all human bodies in an image \cite{Lin}. This is a basic research topic for many visual applications, such as human motion recognition and human-computer interaction. With the recent development of neural networks, many methods perform well using CNN, including stacked hourglass networks \cite{Newell} and multiscale structure-aware networks \cite{Ke1}. %\cite{chen}, \cite{Cao}, \cite{Yang2D}. 
Some researchers have estimated 2D human poses based on heat maps.
Others approaches \cite{Zhou}, \cite{Wang}, \cite{Human3.6M} regressed the heat maps for 3D pose estimation. In addition, Shotton et al. \cite{Shotton} and Junget et al. \cite{Junget} took depth images as the input to process 3D estimation. Some methods are not only applicable in hand pose estimation but also to regress the human body joints, such as the region ensemble network \cite{Guo} and the voxel-to-voxel network \cite{v2v}. Compared to color images, depth images contain 3D information about the distance between the scene object and the camera viewpoint. With this considerable advantage, using 3D information can make a better 3D prediction.

3D interaction devices such as the Intel RealSense, Microsoft Kinect sensor, and
other types of depth/IR cameras have a positive impact on the technology of human-computer interaction (HCI) \cite{Sensor_Kinect}, \cite{Sensor_app}.
Images captured by depth sensors are widely used in robotics and autonomous driving. Point cloud maps, as a form of data structure of depth images, have both geometric positions and intensity information. %With the widespread use of radar and depth cameras in robotics and autonomous driving,
The study of point cloud maps has gradually improved from geometric feature extraction to high-level understanding, such as point cloud segmentation \cite{Sensor_seg} and recognition \cite{pointnet}. Different from ordinary image perception, pose estimation from a point cloud is challenging. First, the data structure of the point cloud is a set of points that are composed of point coordinates in three-dimensional space. These points only provide one-sided geometric information. The limited information cannot be used to extract enough features. Second is the sparsity of the point cloud. When the same object is scanned by different 3D devices in different positions, the order of the three-dimensional points varies widely. Such data are difficult to directly process through an end-to-end model. Third, we need to consider the time and space complexities in performing real-time tasks. Recent networks that directly process each point individually from the point cloud are computationally expensive. Fourth, 3D skeleton joints of most datasets are captured from a depth camera instead of using a Vicon camera to track the markers with high accuracy.

\begin{figure}[htp]
\centering
\includegraphics[scale=0.45]{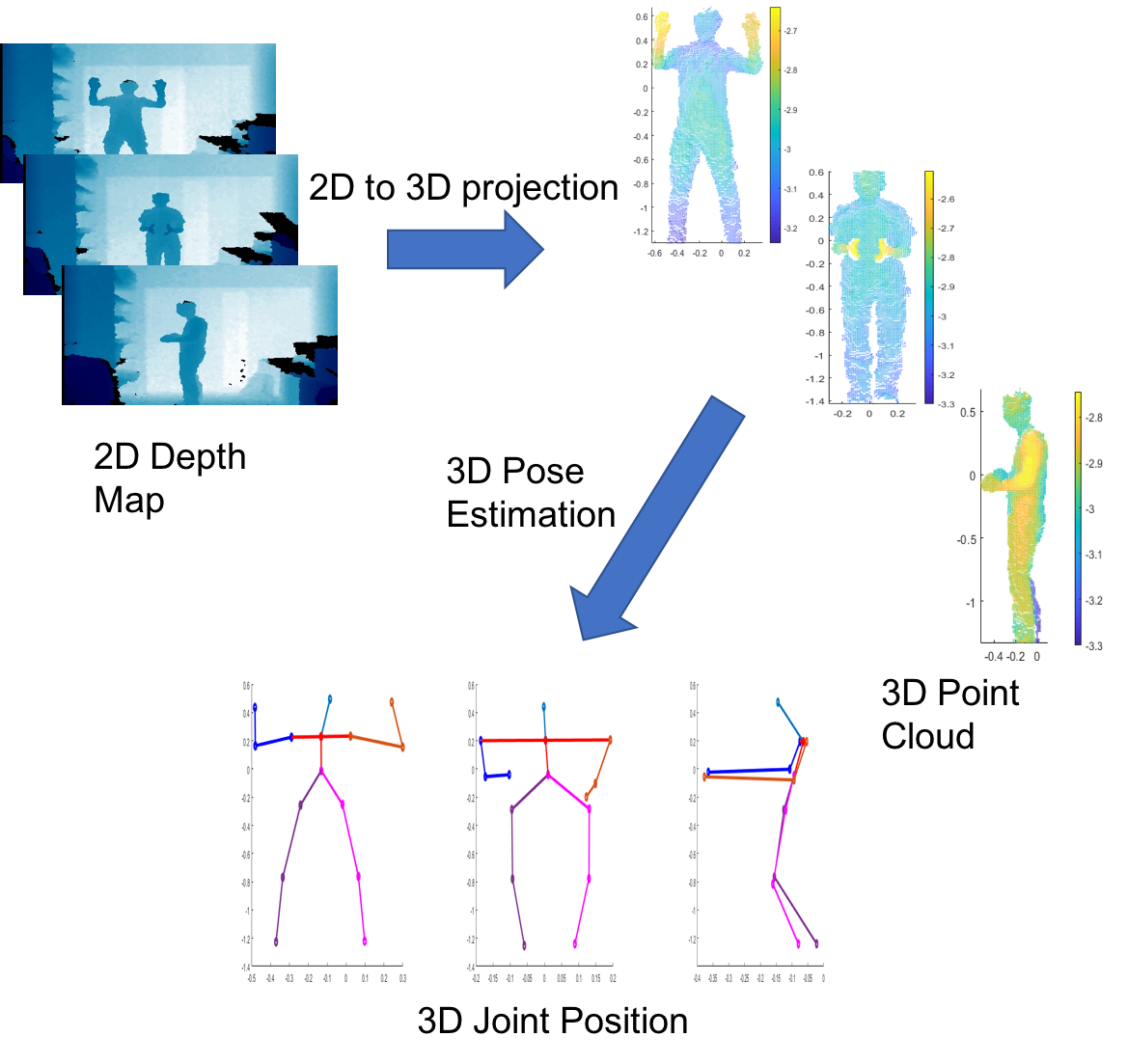}
\caption{ The 3D point cloud has a one-to-one relation with a 3D pose. Our approach is based on point clouds, converting depth images into point clouds before pose estimation.} \label{fig:1}
\end{figure}

What we achieve by converting the depth image into a point cloud is not the actual 3D data, but the 2.5D data from the object surface, which is not suitable for directly processing the point cloud by 2D CNN or 3D CNN. However, the time and space complexities of the 3D CNN grow cubically with the resolution of the input 3D volume. As Figure 1 shows, we aim to learn 3D human key body joint positions directly from the 3D point cloud. PointNet++ \cite{pointnet++} and DGCNN (Dynamic Graph CNN) \cite{DGCNN} have been motivated by the recent works on PointNet \cite{pointnet} that perform both 3D object classification and 3D object segmentation on point clouds directly. 
%{\color{red}Point clouds are fundamentally irregular, and it is not sensitive to the order of the data. This means that the model for processing point cloud data needs to be invariant to different permutations of the data. Qi et al. \cite{pointnet} \cite{pointnet++} proposed PointNet and PointNet++ to directly process the point cloud. In their model, a spatial transform network named T-Net (part of PointNet \cite{pointnet}) has been designed to ensure the invariance of the model to a specific spatial transformation. The points are first aligned by multiplying it with a transformation matrix learned by a spatial transform network. PointNet extracts a global feature for all point cloud data, ignoring local features. PointNet ++ extracts local features on different scales and obtains deep features through a multilayer network structure. PointNet ++ will first sample the point cloud (sampling) and divide the area (grouping), and use the basic PointNet network for feature extraction in each small area, and iterate continuously. For a convenient local feature extracting, Wang et al. [15] proposed a EdgeConv layer in their DGCNN model. The EdgeConv layer forms local features by searching neighbor points. Through stacking EdgeConv layers or recycling, global shape information can be extracted. In our model, we directly use T-Net as the spatial transform network to achieves permutation invariance of points and EdgeConv layers to extract and learn the features.}

In this approach, our idea to perform pose estimation is that first, we take the images as input and convert them into 3D point clouds. Depending on the depth thresholding and Euclidean cluster extraction, we extract the person from these 3D points. Points of the human body are normalized by the height and width of this person before being transferred into the deep learning network. 
Our network is designed on a modified dynamic graph CNN model and PointNet, whose output is a low dimensional representation of the 3D key points. This work captures 3D structures of the human body and accurately estimates 3D human poses.

Our contributions can be summarized as follows.
\begin{itemize}
\item To the best of our knowledge, we proposed estimating the human body key joints directly from 3D point clouds based on the network architecture of DGCNN and PointNet for the first time. Unlike other methods that regress the 3D key points from the depth images, we cast the problem of 3D pose estimation from a single depth image to the point clouds. %We transforming a depth map into a 3D point cloud and propose to regress 3D key points directly from the 3D point clouds based on the network architecture of PointNet.

\item We processed the experiment using two existing representative 3D human pose datasets--EVAL\cite{EVAL} and ITOP\cite{Haque}. We compared our approach with other CNN-based \cite{Haque} and RF/RTW-based \cite{Shotton}, \cite{Junget}, \cite{IEF} human pose estimation methods from depth images. Experimental results show that our network for 3D pose estimation has a significantly accurate performance.

%In order to segment the person directly, points of human body are identified by euclidean cluster extraction and thresholding distance to their nearest point when we training. During the real-time testing, we get the rectangle boxes of the human body and corresponding points after tracking the person in the color image by using intel realsense ZR300. 
%to evaluate our method
\end{itemize}

\begin{figure*}[htp]
\centering
\includegraphics[scale=0.52]{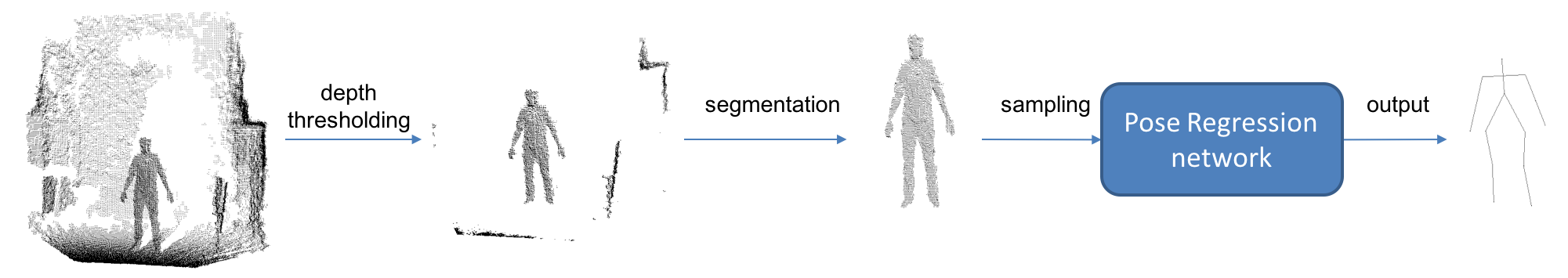}
\caption{The point cloud is converted from the depth image. We define distance thresholds to identify the subject and segment the human body from the background. After sampling the point cloud into the same size, it is fed into the regression network. The output is the 3D coordinates of the skeleton joints.
 } \label{fig:2}
\end{figure*}

\section{Related Work}

\subsection{Depth-based 3D Pose Estimation}
The methods of depth-based 3D pose estimation are divided into two parts: generative and discriminative models. The generative models are similar to the template matching. The human body templates are required to find correspondence between the inputs and templates in generative models. The iterative closest point algorithm \cite{EVAL} is commonly used to track the 3D human body. Different from generative models, discriminative models directly estimate the pose. In the following, we focus on conventional discriminative models.

Most of the discriminative models are based on random forests (RF). Shotton et al. \cite{Shotton} classified body parts from a single depth image based on the random forest classifier and estimated the 3D joint locations. Jung et al. \cite{Junget} used a random tree walk algorithm (RTW) to regress the joint position and reduce the running time. In the field of deep learning, Haque et al. \cite{Haque} proposed a viewpoint-invariant model using CNN and recurrent networks for human pose estimation, while Guo et al. \cite{Guo} introduced a tree-structured region ensemble network for 3D position regression. These deep learning models are based on the image features. In addition, a voxel-to-voxel network (V2V-PoseNet) is proposed in \cite{v2v}, which takes the point cloud as input. For each voxel, the network estimates the likelihood of each body joint. V2V-PoseNet extracts the 3D joint positions from the generated heatmaps.

\subsection{3D Deep Learning For Point Cloud}

A series of PointNet models, including PointNet \cite{pointnet} and PointNet++ \cite{pointnet++} are the recently proposed methods for point cloud classification and segmentation.
%{\color{red}
Point clouds are fundamentally irregular, and it is not sensitive to the order of the data. This means that the model for processing point cloud data needs to be invariant to different permutations of the data. %PointNet and PointNet++ can directly process the point cloud.In their model, 
A spatial transform network named T-Net (part of PointNet \cite{pointnet}) has been designed to ensure the invariance of the model to a specific spatial transformation. The points are first aligned by multiplying it with a transformation matrix learned by a spatial transform network. PointNet models treat each point individually, learning the mapping from 3D to potential features without taking advantage of geometry. A single maxpooling layer is used for all the features of sampled points in PointNet. This operation loses many local features and only maintains global features.
%PointNet is used for a single sampling point, no matter how fine the feature is for a certain sampling point, and the network that integrates all sampling point features only has that maxpooling.
Therefore, the network's ability to extract local information from the model is far less satisfactory than that of the convolutional neural network. 
%{\color{red}
PointNet ++ extracts features on different scales and obtains deep features through a multilayer cascade network. There are three main modules, sampling, grouping and feature learning, for extracting both local and global features. However, PointNet++ still considers individual points in local point sets instead of the relationships between a pair of points.

%The successful development of the convolution neural network (CNN) for image analysis, suggests the value of adapting insight from CNN to the world of point clouds.
 
In contrast to previous networks, Wang et al. \cite{DGCNN} successfully developed a dynamic graph CNN model (DGCNN) to process point clouds based on a thinking way of image analysis using a convolution neural network (CNN). The method is the same as the principle of a convolutional network that considers the set of interconnected pixels instead of a single pixel. EdgeConv, as the main part of DGCNN, takes a center point with its nearest neighbor points as an input to the multilayer perceptron (MLP) layer. %{\color{red}
EdgeConv models the distance and geometric structure between points when constructing the neighborhood. Local features are formed by searching neighbor points, while global features can be extracted through stacking or recycling EdgeConv layers. However, unlike PointNet and PointNet++, not only global features but also local features are considered in EdgeConv layers.

%\textbf{Transfer learning in Deep learning}:Transfer learning is a popular method in computer vision because it allows us to build accurate models in a timesaving way (Rawat & Wang 2017). With transfer learning, instead of starting the learning process from scratch, you start from patterns that have been learned when solving a different problem. This way you leverage previous learnings and avoid starting from scratch. Take it as the deep learning version of Chartres' expression 'standing on the shoulder of giants'.  In computer vision, transfer learning is usually expressed through the use of pre-trained models. A pre-trained model is a model that was trained on a large benchmark dataset to solve a problem similar to the one that we want to solve. Accordingly, due to the computational cost of training such models, it is common practice to import and use models from published literature (e.g. VGG, Inception, MobileNet). A comprehensive review of pre-trained models’ performance on computer vision problems using data from the ImageNet (Deng et al. 2009) challenge is presented by Canziani et al. (2016).

\section{Deep Learning Model}
Our method for pose estimation takes the point clouds converted from the depth images as input, considering this $F$-dimensional point cloud with $N$ points. The point cloud is downsampled to N points, which is defined as a set of vectors $P=\{p_1,p_2,...,p_N\}$. Outputs are a set of 3D key body joint locations $J=\{j_1,...,j_{M}\}$, where $M$ is the number of key body joints and $j_i$ contains $x_i, y_i, z_i$ in the camera coordinate system where $i\in \{1,....,M\}$. The data input to the deep learning model is $(P,J)$. Considering the resolution of the depth image, we set $N$ as $5,000$ and $F$ as $3$. Our model takes these sampled point clouds to regress the 3D body pose by extracting the global and local features. To further improve the results, we modified the DGCNN model. In this section, we introduce our preprocessing and then present our 3D pose estimation method.

\subsection{Preprocess}

First, our approach segments the human body from the background as clearly as possible. Figure 2 shows the process of segmentation and sampling. Since the data are captured by the depth camera, many points with invalid depth values or zero-depth values may be included. We removed all invalid and zero points. Given the depth information, most of the background points can be easily removed by defining depth thresholding. However, only setting the depth thresholding is not enough to extract the human body. Point clouds still include noise and other background objects, which may affect the results. These are mostly due to the photon shot noise and long distance between the human body and the camera. %Noise and background subjects are also showed in Figure 2.

Setting a bounding box and depth thresholding can eliminate many background objects from Figure 2. In a more general case, we can make use of nearest neighbors and implement a clustering technique that is essentially similar to a flood fill algorithm \cite{flood-fill}. The main idea for removing background subjects in point clouds is using the Euclidean cluster extraction filter as shown in Algorithm 1. $Search({n}_i,d_{th})$ is to search for the neighbors of ${n}_i$ with a radius $ r < d_{th}$. The distance thresholding $d_{th}$ is set to $10cm$. $Size(x)$ returns the size of the cluster $x$. Each point from the point cloud is checked whether the distance between it and its neighbors is below thresholding. If the conditions are satisfied, point and its neighbors belong to the same cluster. Since point cloud data provide higher dimensional data, there is considerable information that can be extracted. In our proposed approach, segmentation based on Euclidean distance was performed by removing the small noise cluster.

After the previous steps, considering the different numbers of points, we downsampled the number of original point clouds into the same size. Normalization of the point cloud is executed by the person's height ($b_h$) and width ($b_w$) in this point cloud to address the data of different sizes. Since joint coordinates are in absolute image coordinates, it proves beneficial to normalize them with a bounding box $b$, which includes the person's height ($b_h$) and width ($b_w$). In a trivial case, the box can denote the full image. Such a box is defined by its center. The center in the point cloud is the human body center $b_c$. $b_c$ can be derived from the average of all joints from the ground truth.
As shown in Equations 1 and 2, $NOR(p_i,b)$ is the function for point cloud normalization.
\begin{equation}
\begin{aligned}
NOR(p_i,b)=(p_i-b_c)\begin{bmatrix}
     \frac{1}{b_w} &0 &0 \\    0       & \frac{1}{b_h} &0\\
    0  &0 &1
 \end{bmatrix}
\end{aligned}
\end{equation}
\begin{equation}
\begin{aligned}
b_c=\frac{\sum_{i=0}^{N}p_i}{N}
\end{aligned}
\end{equation}
where $N$ is the number of points from the human body and $p_i$ is the point from point cloud $P$ of the human body.

\begin{figure}[htp]
\centering
\includegraphics[scale=0.3]{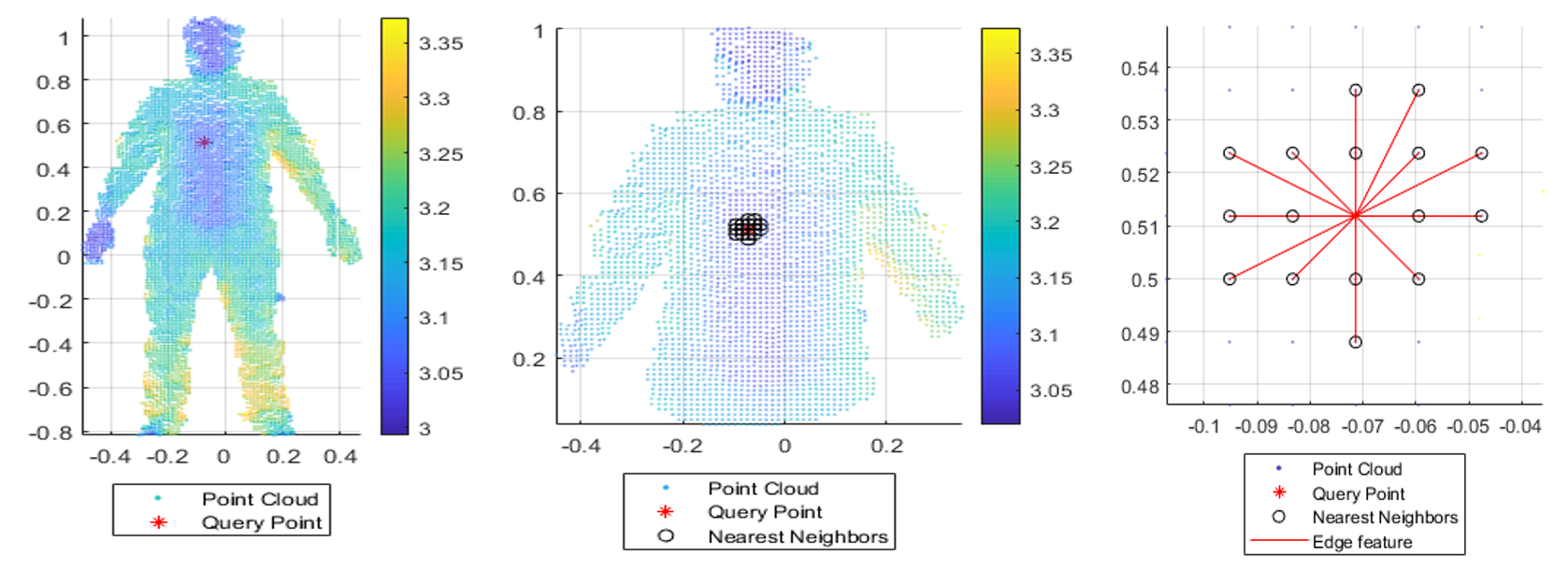}
\caption{An example of edge features from the chest. We define an arbitrary center query point from the chest and its neighborhoods. Depending on the distance, we can find the $K$ neighbor points. The edge features are composed of these $K+1$ points. The point clouds with edge features are the input to the neural network. We set $K$ to $16$ in this work.}
%We define a center point $c_i$ ($c_i \in P$). Depending on the distance, we can Finding the $H$ neighbor points. In this figure, we set $H$ to $4$.
\label{fig:2}
\end{figure}

{
 \begin{algorithm}
 \caption{Euclidean Cluster Extraction}
 \begin{algorithmic}[1]
 \renewcommand{\algorithmicrequire}{\textbf{Input:}}
 \renewcommand{\algorithmicensure}{\textbf{Output:}}
 \REQUIRE Point cloud $P$
 \ENSURE  A point set of human body $C_{body}$ 
 %\\ \textit{Initialisation} :
  \STATE set an empty queue $Q$ and an empty list $C$\;
 %\\ \textit{LOOP Process}
  \FOR {${p}_i \in P$}
  \STATE add ${p}_i$ to $Q$\;
      
    \FOR{${q}_i \in Q$}
      \STATE $Neighbors$=$Search({q}_i,d_{th})$
      \FOR{${n}_i \in Neighbors$}
        \IF{${n}_i$ $\notin$ $Q$ }
         \STATE add ${n}_i$ into  $Q$
         %\STATE remove ${n}_i$ from $P$
        \ENDIF
       \ENDFOR
    \ENDFOR
      \STATE add $Q$ to $C$ 
       \STATE set $Q$ to empty 
  \ENDFOR
  \STATE $C_{body} = Argmax_{x \in C} Size(x)$\;
 \RETURN $C_{body} $ 
 \end{algorithmic} 
 \end{algorithm}
}

\begin{figure*}[htp]
\centering
\includegraphics[scale=0.8]{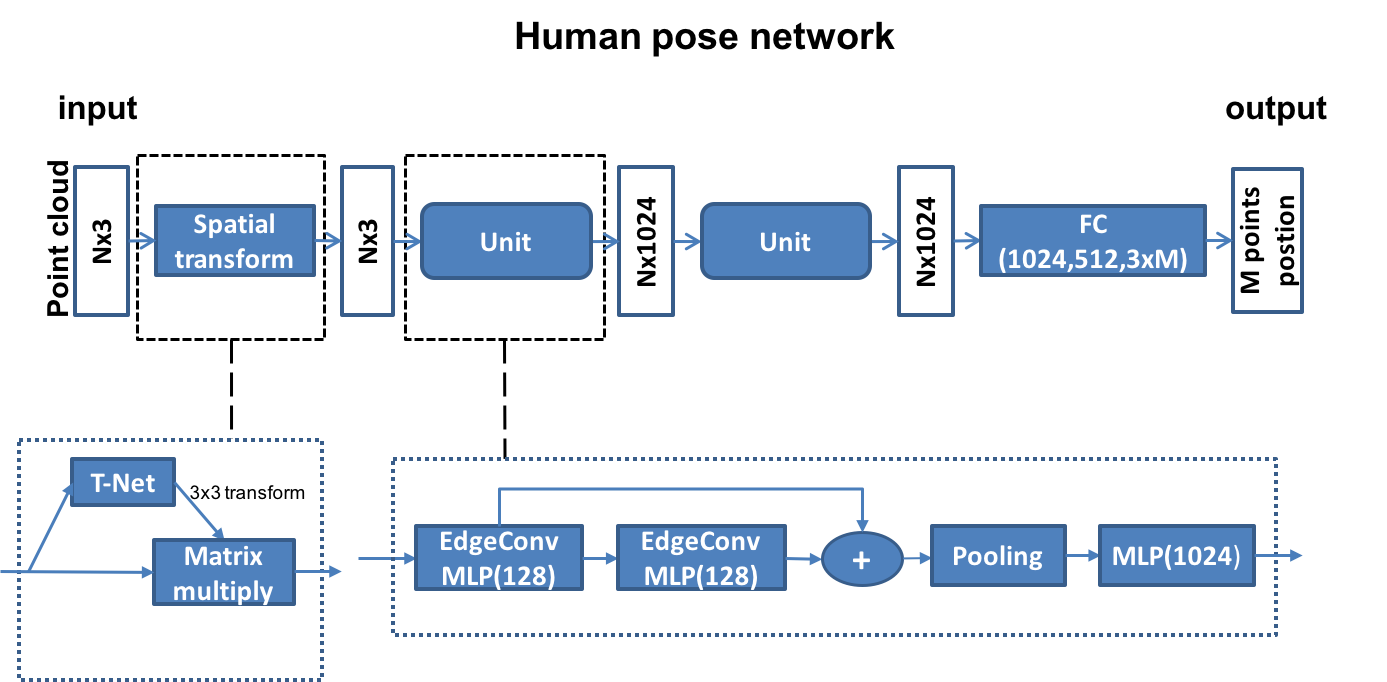}
\caption{The architecture of the pose regression network contains a spatial transform network and two units which are composed of EdgeConv layers. The normalized 3D point clouds are fed into the regression network. The dimension of input is the $N$ $\times$ $3$ point clouds while the dimension of output is $M$ $\times$ $3$. $M$ represents the number of keypoint positions. The normalized 3D point clouds are input to $N$ $\times$ $3$. $\oplus$ represents the process of concatenation. Spatial transform networks are part of PointNet \cite{pointnet}, while EdgeConv is a part of the DGCNN \cite{DGCNN}. FC is the fully connected layer. Our network is trained in an end-to-end manner to extract body features and regress 3D joint locations.
} \label{fig:2}
\end{figure*}

\subsection{Pose Regression Network}

We designed a 3D pose regression network that can directly estimate the 3D coordinates of the pose from the input point clouds by modifying DGCNN and PointNet. %{\color{red}
In our model, we use T-Net as the spatial transform network to achieves permutation invariance of points and stack EdgeConv layers to extract and learn features from the transformed point clouds.
The original DGCNN model is used for classification and segmentation. We modified the architecture of the classification model by changing the middle layers and fully connected layers. The architecture of the modified model is shown in Figure 4. 
Different from the original model, we use one EdgeConv layer with 128 filters instead of three 64 filters. EdgeConv is a basic part of the DGCNN, which captures the local structures of the human body by combining the points and their neighborhoods. In our model, EdgeConv has been used as a main part for feature extraction. As Figure 3 shows, we visualize an edge feature that is considered a local feature. The feature is composed of the $K$ nearest neighbors by calculating the Euclidean distance. EdgeConv takes the local feature as input, considering the coordinates of points and the distance from the domain points as local domain information. Global shape information can be extracted by stacking or recycling EdgeConv layers.

For each point $p_i \in P$, we find its $K$ neighbor points and concatenate the neighbor points with the original point cloud as the edge features, which are calculated in the EdgeConv layers. There are two units in our network, which are composed of EdgeConv layers. The first unit is composed of two EdgeConv (128 filters) layers that are connected with MLP layers (1024 filters). The input to the MLP layers is the concatenation of the outputs from the first two EdgeConv layers. The second unit is the same as the first unit. To regress the joint positions, we modify the last three fully connected layers into $1024$, $512$, $3$ $\times$ $M$. Since we target the single human object, the transform network is considered to be very suitable.
Although the point clouds converted from depth images are ordered, points become an unordered form after preprocessing. %We use a spatial transform network named T-net from PointNet.
The spatial transform network aligns an input point set to canonicalize into a specific space by applying an estimated $3$ $\times$ $3$ transformation matrix. To estimate the $3$ $\times$ $3$ matrix, the network uses the coordinates of each point in the point cloud and the coordinate difference of its $K$ neighbors. After matrix multiplication, the pose network takes the transformed point cloud as input.

\subsection{Network Training}

The process of training is composed of two parts. First, after preprocessing, the point cloud is sampled into a set of $5,000$ points. We trained the whole model with the randomly arranged point cloud as the input in each step.
In this way, the spatial transform network is trained in a good situation. Second, we maintain the weights of the spatial transform network and train the remaining networks. The input to the model is also a set of 5,000 points but not randomly arranged every time. The input to the pose regression network is a set of normalized points $X^{nor}={\{x^{nor}_i\}}^N_{i=1}=\{p^{nor}_i\}^N_{i=1}$, where $p^{nor}_i$ is 3D coordinates of the normalized point and the ground truth $Y^{nor}={\{y^{nor}_i\}}^M_{i=1}={\{j^{nor}_i\}}^{M}_{i=1}$, where $j^{nor}_i$ is the corresponding key body joints after normalization. Given $T$ training samples with normalized point clouds, we minimize the following objective function:

\begin{equation}
\begin{aligned}
\omega^*=\operatorname*{argmin}_w \sum_{t=0}^{T}||Y^{nor}_t-\digamma(X^{nor}_t,\omega)||^2+\lambda||\omega||^2
\end{aligned}
\end{equation}
where $\omega$ denotes the parameters of the pose regression network, $\digamma$ represents the pose regression network, and $\lambda$ is the regularization strength.

%The loss function is mean square error. The mean square error is defined as
%\begin{equation}
%\begin{aligned}
%\centering
%Loss=\frac{\sum_{i=0}^{M}(\hat{j}_i-j_i)^2}{M}
%\end{aligned}
%\end{equation}
%where $\hat{j}$ represent a predicted key joints; $j$ indicates the ground truth; $M$ is the batch size.

%\subsection{Implement details}
%We can design a basic network by changing the last fully connected layer. However, time is seriously considered as the model of full size has about 3 million parameters. We need to cut the size of network and keep the good result. In fact, among the deep learning model, fully connected layers cost lost of time. But changing the fully connected layers to small size can cause the worse results. Finally, we change last three fully connected layers to $1024$,$512$,$3 \times K_{joints}$. After changing the middle (64,64,64) EdgeConv layers into one (128) EdgeConv, we get the totally 2 million parameters and we also get a better result. In order to achieve an ideal situation of the pose regression network. We use the method of transfer learning to train the two parts separately. Training are divided into two steps. First, we train the modified DGCNN model into a best situation since it takes the point cloud as input and output the estimated 3D joint position. Second, we lock the weights of the DGCNN model. We train the refinement network. It takes the output from DGCNN model as the input.

\section{Experiments}
In this section, we evaluate our proposed approach on two public 3D human pose estimation datasets: EVAL \cite{EVAL} and ITOP \cite{Haque}, and then compare our method with state-of-the-art methods.

\begin{table*}[h]
    \setlength{\tabcolsep}{9pt}
    \begin{center}
        % \begin{spacing}
         \caption{Performance comparison of the proposed method with state-of-the-art methods. (RF\cite{Shotton}, RTW \cite{Junget}, IEF \cite{IEF}, VI \cite{Haque})\\ on the ITOP dataset and EVAL dataset}\label{table3}
          
           \begin{tabular}{c|c|c|c|c|c|c|c|c|c|c|c|c|c}
            \hline
             %\multirow{2}{*}{\textbf{Architecture}}
            \textbf{Dataset}&\multicolumn{ 5}{c|}{\textbf{ITOP(front-view)}} &\multicolumn{5}{|c}{\textbf{ITOP(top-view)}}&\multicolumn{3}{|c}{\textbf{EVAL}} \\
            \hline
            \textbf{Body part} &{\textbf{RF}} & {\textbf{RTW}} & {\textbf{IEF}} &{\textbf{VI}} &{\textbf{Ours}} 
            &{\textbf{RF}} & {\textbf{RTW}} & {\textbf{IEF}} &{\textbf{VI}} & {\textbf{Ours}}
             &{\textbf{RTW}} &{\textbf{VI}}& {\textbf{Ours}}\\
            \hline
           %\multirow{2}*{Encoder-Decoder} 
            \textbf{Head} 
            
            &63.8 &97.8 &96.2 &\textbf{98.1}  &96.73
            
            &95.4 &\textbf{98.4} &83.8 &98.1  &96.13

            & 90.9 &\textbf{93.9} &88.93
            \\
           
           \textbf{Neck/Chest} 
           
           &86.4 &95.8 &85.2 &97.5  & \textbf{98.05}
           
           &\textbf{98.5} &82.2 &50.0 &97.6 &97.61
           
            &87.4 &94.7 &\textbf{96.87}
           \\
           
           \textbf{Shoulders} 
           
           &83.3 &94.1 &77.2 &\textbf{96.5} &94.38
           
           &89.0 &91.8 &67.3 &\textbf{96.1} &93.08
           
           &\textbf{87.8} &87.0 &86.14
           \\
           
            \textbf{Elbows}
            
            &73.2 &\textbf{77.9} &45.4 &73.3  &73.67
            
            &57.4 &80.1 &40.2 &\textbf{86.2}  &70.83
           
           &27.5 &45.5 &\textbf{75.11}
           \\
          
            \textbf{Hands}
            &51.3 &\textbf{70.5} &30.9 &68.7 &54.95
            
            &49.1 &76.9 &39.0 &\textbf{85.5} &48.41
            
           &32.3 &39.6 &\textbf{63.07}
           
           \\
            
            \textbf{Torso}
            
            &65.0 &93.8 &84.7 &85.6 &\textbf{98.35}

            &80.5 &68.2 &30.5 &72.9 &\textbf{95.58}
            
            &\textbf{--}&\textbf{--} &\textbf{--}
           \\
           
            \textbf{Hips} &50.8 &80.3 &83.5 &72.0 &\textbf{91.77}
            
            &20.0 &55.7 &38.9 &61.2 &\textbf{84.50}
           &\textbf{--}&\textbf{--} &83.20
           
           \\
            
            \textbf{Knees}
            &65.7 &68.8 &81.8 &69.0 &\textbf{90.74}
            
            &2.6 &53.9 &54.0 &51.6  &\textbf{79.19}
            &83.4 &\textbf{86.0} &82.68
           \\
            
            \textbf{Feet}
            
            &61.3 &68.4 &80.9 &60.8   &\textbf{86.30}
            
            &0.0 &28.7 &62.4 &51.5  &\textbf{67.76}

            &90.0 &\textbf{92.3} &82.94
            
           \\
            \hline
            
               \textbf{Upper Body}&70.7 &\textbf{84.8} &61.0 &84.0  &80.10
             
              &73.1 &84.8 &51.7 &\textbf{91.4}  &77.30
              &59.2 &73.8 & \textbf{79.31}

           \\
           
            \textbf{Lower Body}&59.3 &72.5 &82.1 &67.3  &\textbf{89.60}
            
            &7.5 &46.1 &53.3 &54.7  & \textbf{77.15}
            
            &86.7  &\textbf{89.2} &82.94
            
           \\

             \textbf{Mean}&65.8 &80.5 &71.0 &77.4 &\textbf{85.11}
             
             &47.4 &68.2 &51.2 &75.5 &\textbf{78.46}
             &68.3 &74.1 &\textbf{80.86}
           \\
            \hline
            
    \end{tabular}
         
        % \end{spacing}
    \end{center}
\end{table*}

\subsection{Implementation Details}
All experiments were performed on a computer with one Intel Core i7-9700K CPU, n Nvidia GTX1080ti GPU with 12 GB memory, and 64 GB of RAM. The pose regression network and preprocessing were implemented by the TensorFlow framework and point cloud library (PCL). The optimizations of both training steps were Adam \cite{Adam}. For the first step to train the spatial transform network, the initial learning rate was $0.001$, and decay rates ($\beta_1$, $\beta_2$) were separately $0.9$ and $0.999$. Epsilon of Adam was $1e-08$. The learning rate was divided by 10 after 50 epochs. After approximately 80 epochs, we stopped the first training and started the second training step. For the second step of training, the initial learning rate was $0.00001$. Regularization strength was set to $0.0005$. The number of neighbor points $K$ was 16. All MLP and fully connected layers included the ReLU activation function, except the last layer. We trained the network with a batch size of 4. For a more in-depth training of the spatial transform network, we randomly sampled the data in each training step.
When we performed the evaluation, the estimated 3D key body joint locations were reconstructed from the network outputs:
\begin{equation}
\begin{aligned}
J=\digamma(X^{nor}_t,\omega^*)
 \begin{bmatrix}
   b_w &0 &0 \\    0       & 
   b_h &0\\
    0  &0 &1
 \end{bmatrix}+b_c
\end{aligned}
\end{equation}

\subsection{ Data Preparation and Evaluation Metrics}

\begin{figure}[htp]
\centering
\includegraphics[scale=0.285]{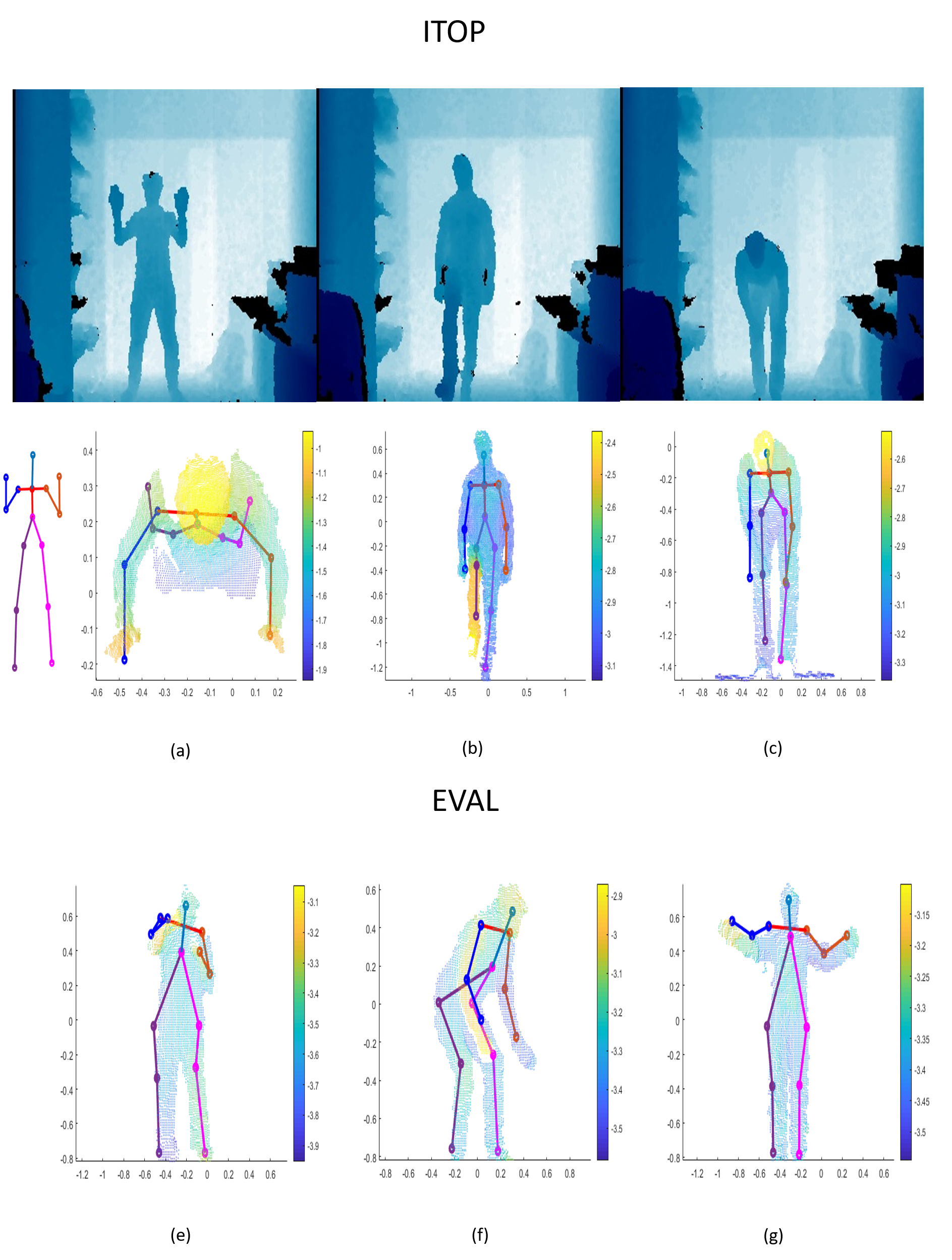}
\caption{An example of ITOP and EVAL dataset. There are $15$ key joints in the ITOP dataset, which are head, neck, torso, shoulders, elbows, hands, hips, knees and feet. There are $14$ key joints in the EVAL dataset, which are head, chest, shoulders, elbows, hands, hips, knees and feet.}
%We define a center point $c_i$ ($c_i \in P$). Depending on the distance, we can Finding the $H$ neighbor points. In this figure, we set $H$ to $4$.
\label{fig:5}
\end{figure}

We evaluated the performance of our point cloud-based human pose regression network on the EVAL and ITOP datasets. Examples of ITOP and EVAL datasets are shown in Figure 5. Figure 5(a) represents the top-view of the ITOP dataset, while (b) and (c) represent the front-view of the ITOP dataset. Depending on the different datasets, we estimate the main $14$ key body joints on the EVAL dataset and estimate $15$ on the ITOP dataset. %{\color{red}
To produce the most convincing test results and verify the generalization ability of our model, all training and testing data are divided by the subjects.

The ITOP 3D human pose estimation dataset consists of two angles of view--front-view and top-view tracks. Each track contains a list of approximately 40 k training and 4 k testing depth images.
The resolution of the image is 320 $\times$ 240 (width $\times$ height). This dataset consists of 20 actors who perform 15 sequences each and is recorded by two Asus Xtion Pro cameras. The ground truth of this dataset is the 3D coordinates of 15 body joints, including head, neck, shoulders, elbows, hands, torso, hips, knees, and feet. The 3D joint position is directly from the camera interface. After we removed invalid data, the dataset has approximately 17k and 4k depth data separately for training and testing. We take 3k of the training data for validation. Both front-view and top-view depth data with their corresponding joint data are used for evaluation.

The EVAL 3D human pose estimation dataset has 3 subjects: 1 female and 2 males. It contains approximately 10 k frames. Each frame is combined with the Vicon data of 30 markers. The resolution of the image is 320 $\times$ 240 (width $\times$ height). The dataset is divided into three parts (training, validation, and testing). We remove some invalid data that are out of the Vicon data. Finally, we have 8,158 front-facing depth images. We choose one subject for the test (approximately 2k frames) and the other two subjects (approximately 5k frames) for training, retaining approximately 800 frames as a validation dataset. The ground truth of this dataset is the 3D coordinates of 14 body joints, including chest, hips, shoulders, elbows, knees, feet, head, hands. Each joint is set up with different Vicon markers, chest with 1 marker, hip with 2 markers, shoulder with 2 markers, elbow with 2 markers, knee with 3 markers, feet with 3 markers, head with 1 marker, and hand with 3 markers. We select the average position of markers for each joint. Different from the ITOP dataset, the coordinates of joints are calculated from the Vicon system and markers. These joints are of high accuracy. %As Figure 5 shows, the joints from the EVAL dataset are more accurate than ITOP dataset due to the vicon system.

\begin{figure*}[htp]
\centering
\includegraphics[scale=0.3]{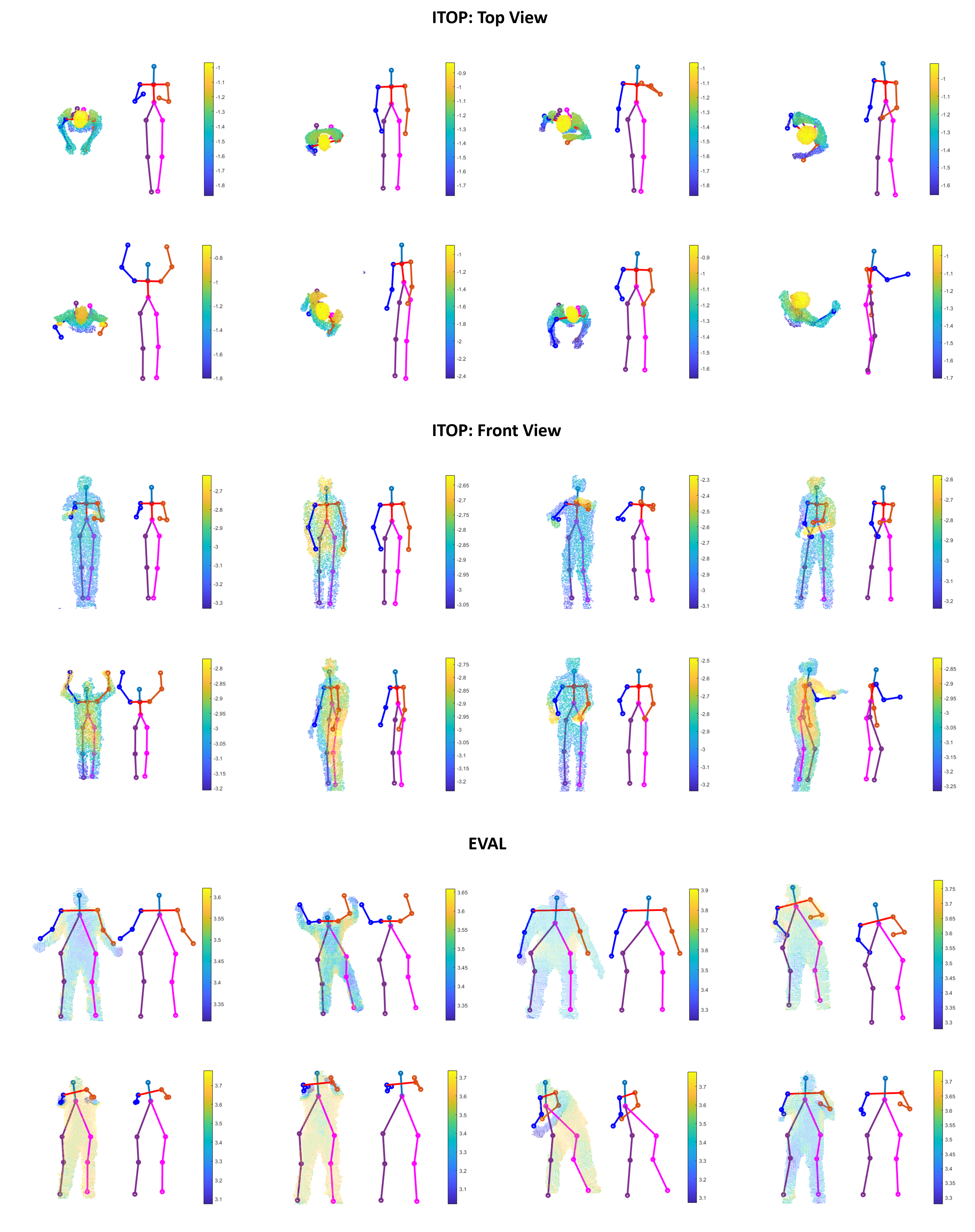}
\caption{Qualitative results of our pose net on EVAL and ITOP (both front-view and top-view) datasets. A segmented point cloud of a human is shown with its corresponding joints. The motions in ITOP are turning around, standing, raising hands and boxing, while the motions in EVAL are standing, shaking, and raising hand.}
%We define a center point $c_i$ ($c_i \in P$). Depending on the distance, we can Finding the $H$ neighbor points. In this figure, we set $H$ to $4$.
\label{fig:5}
\end{figure*}

%In the experiment, we resize the point cloud into $5000$ by random sampling.

We use the mean average precision (mAP) which is defined as the average precision for all human body parts. We set a $10cm$ rule following \cite{Junget}, \cite{Haque} to show the result of each kind of joint, which means there is a successful detection when the predicted joints are less than $10cm$ from the ground truth. Here are the evaluation metrics.
\begin{equation}
AP(x,y)= \begin{cases}1, & \text{if } distance(x,y) < 10 cm\\
    0,              & \text{otherwise} 
    \end{cases}
\end{equation}
\begin{equation}
\begin{aligned}
mAP =\frac{\sum_{i=0}^{M}AP(J_i,GT_i)}{M}
\end{aligned}
\end{equation}
where $distance(x,y)$ denotes the Euclidean distance between $x$ and $y$ in 3D space; $J$ and $GT$ are predicted joints and ground truth separately. $M$ is the total number of joints.

\subsection{Comparison with State-of-The-Art Methods}

We compared our proposed approach with state-of-the-art methods, which include random forest (RF)\cite{Shotton}, the random tree walk algorithm (RTW) \cite{Junget}, iterative error feedback (IEF) \cite{IEF}, and the viewpoint-invariant feature-based method (VI) \cite{Haque} on the ITOP dataset. Next, we conducted training on the EVAL dataset and compared our approach with RTW and VI.
The mAP of the upper body and lower body is provided in Table I. The upper part of the body includes head, neck/chest, shoulders, elbows, and hands, while the lower body includes hips, knees, and feet. The qualitative results of our approach on the ITOP front-view, ITOP top-view, and EVAL datasets are shown in Figure 6. The motions in Figure 6 are turning around, standing, raising hands, boxing and shaking.

Most of the methods mentioned above performed well on the main parts of the human body, which are head, neck/chest, shoulders, and torso. These main parts provide the richest depth information. Predicting the 3D position of feet and hands is the hardest task in 3D human pose estimation since the proportion of hands and feet to the body is relatively small. In contrast to previous methods, our approach has excellent performance on the joints of the feet and knees. Some kinds of joints, such as knees, hands, elbows, hips, and feet, may be invisible from the view of the depth camera depending on the human motions, which results in deviation of the estimation. %As Figure 5 (b) shows, the right foot is marked on the left foot. The model may wrongly estimate the left hand as the right hand or left leg as the right leg.
However, we still perceive from Table I that our method corresponds to this problem well.

For the front-view part of the ITOP dataset, RTW has the best score on estimating both upper body joints with a mAP of $84.8$. Our predictions of other parts except hands are not much different from the state-of-the-art approaches because parts of the edge of the body, such as the hand, always convey limited depth information. The scores of RF, RTW, IEF, and VI are lower than ours in predicting feet and knees. Our approach has a high score with a mAP of $89.60$ on predicting the lower body joints. From Table I, our method performs better than the other methods in predicting the whole body with a mAP of $85.11$.

For the top-view of the ITOP dataset, the view angle of the collected dataset results in a considerable loss of depth information. The top-view of the ITOP dataset suffers more serious depth information loss than the front-view and EVAL. Figure 5 (a) can be taken as a representative example of such a situation. Therefore, among all experiments, the results from the top-view part of the ITOP dataset are the least accurate. Although VI has the best score in the assessment of the upper body with a mAP of $91.4$, our human pose net has a score with a mAP of $77.15$, which is higher than other approaches. On estimating the whole body joints, our approach has a good mAP score performance of $78.46$, which is higher than the second-ranking method VI. From the results of our approach, the estimation accuracy of the upper body part equals the lower part, and the main difference exists in the prediction of the hand part.

Although both ITOP and EVAL have front-view data, the EVAL dataset has more complex motion than ITOP. In the experiments on the EVAL dataset, VI achieved a higher score on the evaluation of the lower body with a mAP of $89.2$. For the whole body, our approach was the first one with the whole body mAP of $80.86$, which was much higher than the others (RTW and VI). This comes from the accurate estimations in the hand and elbow joints. %\textbf{Our approach requires 1.5 seconds per frame.} 

In our experiment, the most challenging joints are hand and elbow. The mean average precision from both datasets are below 75. The main reason is the lack of enough depth information. The motion range of both of the two joints is large, leading that the two parts beyond the capturing scope of the camera. Moreover, these two joints are easily occluded by other body parts from the view of the camera. Once these two parts are partially occluded, it results in that the extracted body parts in the point cloud image are not connected with each other and cause the depth information to be filtered out in the processing of segmentation.

\subsection{Discussion}
The application scenarios of our proposed approach are wide, including rehabilitation training, exercise coaching, sport player tracking, etc. In the rehabilitation training, most of the environments are controlled structured environments and the designed training is typically a slow tracking movement for a single patient. In this situation, our proposed approach well suits the task. In the exercise coaching, the movement can be fast which needs our proposed approach to be efficient enough to track the movement. In this case, we can tailor the model to be a light-weight simplified one by eliminating the neural links with small weights and increasing the number of sharing weights between clustered links. In the sport player tracking, the tasks are typically multi-player tracking which often involves occlusions. To overcome the occlusion issue, we can use maximum likelihood principle or posterior estimation to make correct joint association and pose estimation by fully utilizing the prior knowledge of the constraint between adjacent joints. More specifically, the occlusion can be modeled as a crossing of two segments based on computational geometry principle. Different combinations of movement sequences are then compared in the sense of probability by taking human movement restrictions into account \cite{Bowen}.

%However, we think the main impact is whether it can effectively extract enough depth information of the human body from the depth camera. In a complex environment, segmenting surrounding objects is the biggest impact on our method. If the user is standing too close or too far to the camera, an invalid interception of human information may happen, which leads to a decline in the accuracy of the prediction.

\section{Conclusion}
We proposed a human pose model that estimates human pose from point clouds with a single depth image. A 2.5D depth image is converted to a 3D point cloud and processes it in the directly modified dynamic graph CNN network and PointNet to regress the 3D positions of joints. %Although we have shown that our model achieves good performance on the two different public depth-based pose estimation dataset, there is still lots of fields to be developed.
To handle the influence of background objects, we segment the point cloud of the human body using Euclidean cluster extraction. After being normalized by its height and width, the point cloud is fed into our modified dynamic graph CNN network. Experimental results compared to other state-of-the-art methods on two public depth-based datasets show that our method performs well for 3D human pose estimation.
Our future work will be further improving accuracy for fast and occluded movements. Moreover, we will testify our approach in the task-oriented datasets, such as rehabilitation, excise coaching, and sport team players' tracking \cite{KIMORE, negin2013decision}.

%Last but not least, we need a bigger public front-facing human motion dataset with more accuracy joints data. %We hope our work to provide a new way of accurate 3D human pose estimation.

% \nocite{*}
\bibliographystyle{ieeetr}
\bibliography{reference}
% \vskip -2\baselineskip plus -1fil

%\appendices

%\section{}

%\section*{Acknowledgment}


%\ifCLASSOPTIONcaptionsoff
%  \newpage
%\fi

%\begin{thebibliography}{1}


%\end{thebibliography}

\begin{IEEEbiography} [{\includegraphics[width=1in,height=1.25in,clip,keepaspectratio]{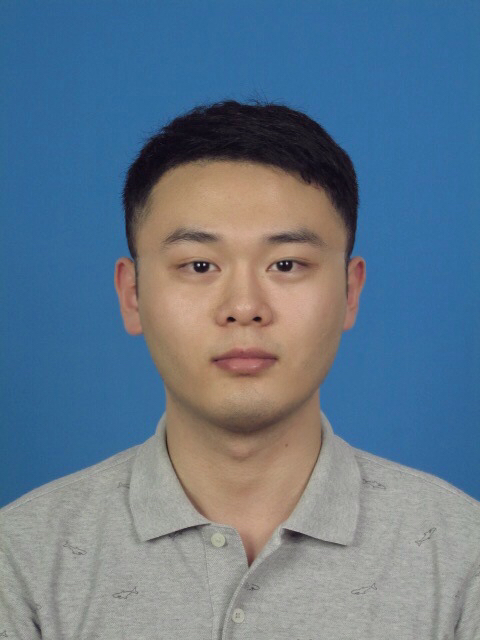}}]{Yufan Zhou} received the B.Eng. degree in Railway Traffic Signaling and Control from Southwest Jiaotong University, China, in 2017. He is currently pursuing the M.A.Sc. degree in electrical and computer engineering at the University of Ottawa. His research interests include artificial intelligence and multimedia.
\end{IEEEbiography}

\begin{IEEEbiography}[{\includegraphics[width=1in,height=1.25in,clip,keepaspectratio]{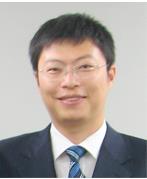}}]{Haiwei Dong} (M'12-SM'16) received the Dr.Eng. degree in computer science and systems engineering from Kobe University, Kobe, Japan, and the M.Eng. degree in control theory and control engineering from Shanghai Jiao Tong University, Shanghai, China, in 2010 and 2008, respectively. He was a Research Scientist with the University of Ottawa, Ottawa, ON, Canada; a Postdoctoral Fellow with New York University, New York City, NY, USA; a Research Associate with the University of Toronto, Toronto, ON, Canada; a Research Fellow (PD) with the Japan Society for the Promotion of Science, Tokyo, Japan. He is currently a Principal Engineer with Huawei Technologies Canada, Ottawa, ON, Canada and a licensed Professional Engineer in Ontario. His research interests include artificial intelligence, robotics, and multimedia.
\end{IEEEbiography}

\begin{IEEEbiography}[{\includegraphics[width=1in,height=1.25in,clip,keepaspectratio]{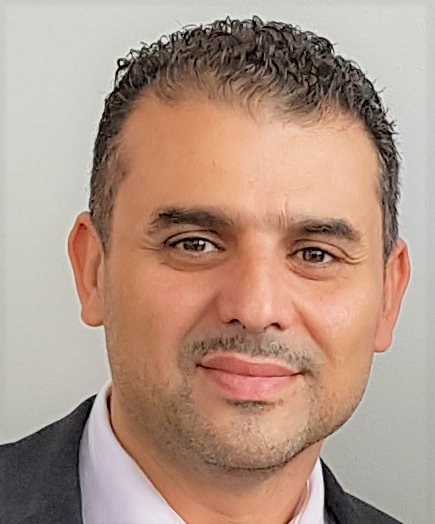}}]{Abdulmotaleb El Saddik} (M'01-SM'04-F'09) is Distinguished University Professor and University Research Chair in the School of Electrical Engineering and Computer Science at the University of Ottawa. His research focus is on the establishment of Digital Twins to facilitates the well-being of citizens using AI, IoT, AR/VR and 5G, hence allowing people to interact in real-time with one another as well as with their smart digital representation. He has co-authored 10 books and more than 550 publications and chaired more than 50 conferences and workshop. He has received research grants and contracts totaling more than \$20 M. He has supervised more than 120 researchers and received several international awards, among others, are ACM Distinguished Scientist, Fellow of the Engineering Institute of Canada, Fellow of the Canadian Academy of Engineers and Fellow of IEEE, IEEE I\&M Technical Achievement Award, IEEE Canada C.C. Gotlieb (Computer) Medal and A.G.L. McNaughton Gold Medal for important contributions to the field of computer engineering and science.
\end{IEEEbiography}

\end{document}